\documentclass{article}

     \usepackage[nonatbib, final]{neurips_2020}

\usepackage[utf8]{inputenc} 
\usepackage[T1]{fontenc}    
\usepackage{hyperref}       
\usepackage{url}            
\usepackage{booktabs}       
\usepackage{amsfonts}       
\usepackage{nicefrac}       
\usepackage{microtype}      
\usepackage{xcolor}

\usepackage{algorithm}
\usepackage{algorithmic}
\usepackage{booktabs}
\usepackage{graphicx}

\usepackage{mwe}

\newlength{\tempdima}
\newcommand{\rowname}[1]
{\rotatebox{90}{\makebox[\tempdima][c]{#1}}}

\newcounter{subfigure}[figure]
\renewcommand{\thesubfigure}{\alph{subfigure}}
\newcommand{\mycaption}[1]
{\refstepcounter{subfigure}\textbf{(\thesubfigure) }{\ignorespaces #1}}

\title{Implicit Rank-Minimizing Autoencoder}

\author{%
Li Jing\\
Facebook AI Research\\
New York\\
\texttt{ljng@fb.com}
\And 
Jure Zbontar\\
Facebook AI Research\\
New York\\
\texttt{jzb@fb.com}
\And 
Yann LeCun\\
Facebook AI Research\\
New York \\
\texttt{yann@fb.com} \\
}

\begin{document}

\maketitle

\begin{abstract}
An important component of autoencoders is the method by which the information capacity of the latent representation is minimized or limited. In this work, the rank of the covariance matrix of the codes is implicitly minimized by relying on the fact that gradient descent learning in multi-layer linear networks leads to minimum-rank solutions. By inserting a number of extra linear layers between the encoder and the decoder, the system spontaneously learns representations with a low effective dimension. The model, dubbed Implicit Rank-Minimizing Autoencoder (IRMAE), is simple, deterministic, and learns compact latent spaces. We demonstrate the validity of the method on several image generation and representation learning tasks.
\end{abstract}

\section{Introduction}
Optimizing a {\em linear} multi-layer neural network through gradient descent leads to a low-rank solution. This phenomenon is known as implicit regularization and has been extensively studied under the context of matrix factorization \cite{Gunasekar2018ImplicitRI, Arora2019ImplicitRI, Razin2020ImplicitRI}, linear regression \cite{Saxe2019AMT, Gidel2019ImplicitRO}, logistic regression \cite{Soudry2018TheIB}, and linear convolutional neural networks \cite{Gunasekar2018ImplicitBO}. The main goal of these prior works were to understand the generalization ability of deep neural networks. By contrast, the goal of the present work is to design an architecture that takes advantage of this phenomenon to improve the quality of learned representations.

Learning good representations remains a core issue in AI \cite{Bengio2013RepresentationLA}. Representations learned in a self-supervised (or unsupervised) manner can be used for downstream tasks such as generation and classification. Autoencoders (AE) are a popular class of method for learning representations without requiring labeled data. The internal representation of an AE must have a limited information capacity to prevent the AE from learning a trivial identity function. Variants of AEs differ by how they perform this limitation. Bottleneck AE (sometimes called "Diabolo networks") simply use low-dimensional codes~\cite{rhw-1986}, noisy AE, such as variational AE add noise to the codes while limiting the variance of their distribution~\cite{doi-nips-2004,Kingma2014AutoEncodingVB}, quantizing AE (such as VQ-VAE) quantize the codes into discrete clusters~\cite{Oord2017NeuralDR}, sparse AE impose a sparsity penalty on the code~\cite{Ng2000SparseAE,ranzato-nips-07}, contracting and saturating AE minimize the curvature of the network function in directions outside the data manifold~\cite{Rifai2011ContractiveAE,goroshin-lecun-iclr-13}, and denoising AE are trained to produce large reconstruction error for corrupted samples~\cite{Vincent2008ExtractingAC}.

In this work, we propose a new method to implicitly minimize the rank/dimensionality of the latent code of an autoencoder. We call this model Implicit Rank-Minimizing Autoencoder (IRMAE). This method consists in inserting extra linear layers between the encoder and the decoder of a standard autoencoder. This additional linear network is trained jointly with the rest of the autoencoder through classical backpropagation. As a result, the system spontaneously learns representations with a low effective dimensionality. Like other regularization methods, this extra linear neural network does not appear at inference time as the linear matrices collapse into one. Thus, the encoder and the decoder architecture of the model is identical to the original model. In practice, we fold the collapsed linear matrices into the last layer of the encoder at inference time.

We empirically demonstrate IRMAE's regularization behavior through a synthetic dataset and show that it learns good representation with a much smaller latent dimension. Then we demonstrate superior representation learning performance of our method against a standard deterministic autoencoder and comparable performance to a variational autoencoder on MNIST dataset and CelebA dataset through a variety of generative tasks, including interpolation, sample generation from noise, PCA interpolation in low dimension, and a downstream classification task. We also conducted an ablation study to verify that the advantage of implicit regularization comes from gradient descent learning dynamics.

We summarize our contributions as follows:
\begin{itemize}
    \item We proposed a method of inserting extra linear layers in deep neural networks for rank regularization;
    \item We proposed a simple, deterministic rank-minimization autoencoder that learns low-dimensional representation;
    \item We demonstrated a superior performance of our method compared to a standard deterministic autoencoder and a variational autoencoder on a variety of generative and downstream classification tasks.
\end{itemize}

\section{Related Work}

The implicit regularization provided by gradient descent optimization is widely believed to be one of the keys to deep neural networks' generalization ability. Many works focusing on linear cases are trying to study this behavior empirically and theoretically. Soudry et al. \cite{Soudry2018TheIB} show that implicit bias helps to learn logistic regression. Saxe et al. \cite{Saxe2019AMT} study a 2-layer linear regression and theoretically demonstrated that continuous gradient descent could lead to a low-rank solution. Gidel et al. \cite{Gidel2019ImplicitRO} extend such theory to a discrete case for linear regression problems. In the field of matrix factorization, Gunasekar et al. \cite{Gunasekar2018ImplicitRI} theoretically prove that gradient descent can derive minimal nuclear norm solution. Arora et al. \cite{Arora2019ImplicitRI} extend this concept to the deep linear network case by theoretically and empirically demonstrating that a deep linear network can derive low-rank solutions. Gunasekar et al. \cite{Gunasekar2018ImplicitBO} prove that gradient descent has a regularization effect in linear convolutional networks. All these works are trying to understand why gradient descent can help generalization in existing approaches. On the contrary, we take advantage of this phenomenon to develop better algorithms. Also, the current implicit regularization study requires a small gradient and vanishing initialization, while our method is more general and can be used with complicated optimizers such as Adam \cite{Kingma2015AdamAM} and allow  combination with more complicated components.

Autoencoders are popular for representation learning. It is important to limit the latent capacity as the data are embedded in a lower-dimensional space. A big family of them are based on variational autoencoders \cite{Kingma2014AutoEncodingVB} such as beta-VAE \cite{Higgins2017betaVAELB}. These methods tend to generate blurry images due to its intrinsic probabilistic nature.  On the other hand, a naive deterministic autoencoder is considered a failure in generative tasks and has ``holes'' in its latent space, due to the absence of explicit constraint on the latent distribution. Many methods with deterministic autoencoder are proposed to solve this problem, such as RAE \cite{Ghosh2020FromVT}, WAE \cite{Tolstikhin2018WassersteinA}, VQ-VAE \cite{Oord2017NeuralDR}.

\section{Implicit Rank-Minimizing Autoencoder}

\begin{figure}[h!]
    \centering
    \includegraphics[width=0.85\textwidth]{ 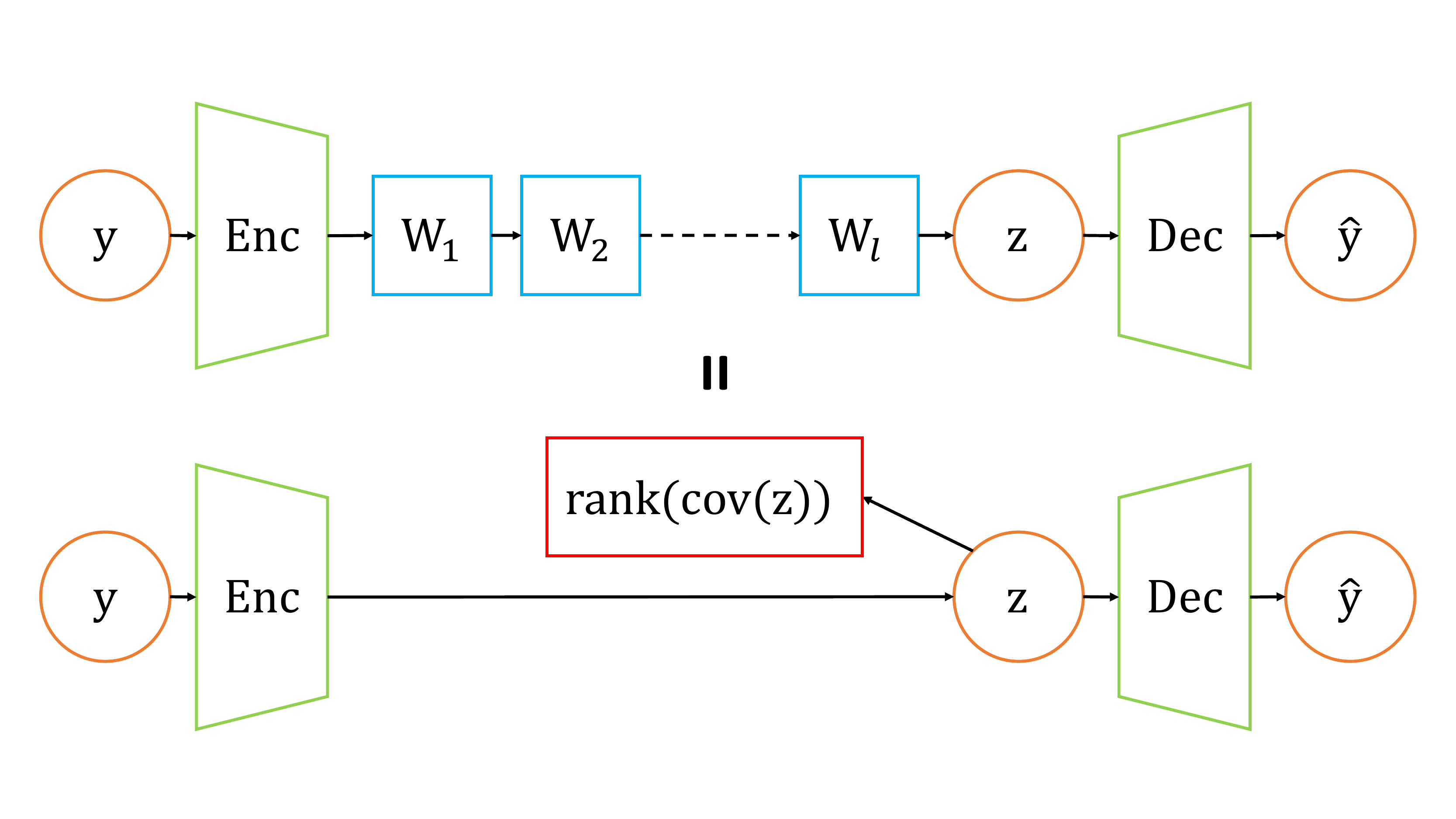}
    \caption{Implicit rank-minimizing autoencoder: a deterministic autoencoder with implicit regularization. The linear matrices that form a linear neural network between the encoder and the decoder are all square matrices. The effect of these matrices is to penalize the rank of the code variable. These matrices are equivalent to a single linear layer at inference time, and thus they do not change the capacity of the autoencoder. In practice, they are absorbed into the last layer of the encoder.}
    \label{fig:model}
\end{figure}

We denote by $\mathcal{E}()$ and $\mathcal{D}()$ the encoder and decoder of a deterministic autoencoder, respectively. The latent variable $z\in \mathbb{R}^d$ is determined by $\mathcal{E}(y)$. Encoder and decoder are classically trained by jointly minimizing the $L_2$ reconstruction loss $L_{AE} = ||y-\mathcal{D}(\mathcal{E}(y))||_2^2$.
Without any constraint on the latent space, a simple deterministic autoencoder will typically learn a non-Gaussian latent space with ``holes'' and hence does not generate good samples.

Implicit rank-minimizing autoencoder consists in adding extra linear matrices $W_1, W_2, \cdots, W_l$ between the encoder and decoder, where $W_i \in \mathbb{R}^{d\times d}$ are randomly initialized. The corresponding diagram is shown in Figure \ref{fig:model}. All $W_i$ matrices are trained jointly with the encoder and the decoder. Hence, the reconstruction loss is represented as
\begin{equation}
L=||y-\mathcal{D}(W_l\cdots W_2W_1\mathcal{E}(y))||_2^2
\end{equation}

During training, these matrices encourage latent variables to use a lower number of dimensions and effectively minimize the rank of the covariance matrix of the latent space. Thus, one can amplify the regularization effect by adding more $W_i$ matrices between the encoder and the decoder. Also, we do not use special initialization of each $W_i$, and it works with more optimizers such as Adam \cite{Kingma2015AdamAM}.

During inference, all $W_i$ matrices can be ``absorbed'' into the encoder as all the linear matrices collapse, as linear matrix multiplication is associative. Therefore, we can directly use this linearly modified decoder for generative tasks; we can also directly use the encoder for downstream tasks such as classification.

\section{Experiment}

In this section, we empirically evaluate the proposed IRMAE model. We first verify the regularization effect through a synthetic task. We then demonstrate that IRMAE generates higher quality images compared to a baseline AE. IRMAE shows comparable performance to VAE. Lastly, we demonstrate IRMAE's superior performance on downstream classification tasks.

Throughout all the experiments, we demonstrate the latent dimension by plotting the normalized singular values. Each plot in Figures \ref{fig:synthetic} and \ref{fig:sv} depicts singular values (sorted from large to small) of the covariance matrix of the latent variables $z$ corresponding to examples in the validation set. The plots are normalized by dividing each singular value by the largest singular value of the covariance matrix. Therefore, the dimension of latent space can be interpreted as the number of nonzero singular values.

\subsection{Verification with Known Intrinsic Dimension}

We verify the regularization behavior of IRMAE via a synthetic shape dataset. Each example is a 32x32 RGB image with a random-color, random-sized square or circle, located at a random position. Hence, the data has a known intrinsic dimensionality of 7 (3 for color, 2 for coordinate, 1 for size, 1 for shape).  

\begin{figure}[h!]
    \centering
    \includegraphics[width=0.5\textwidth]{ 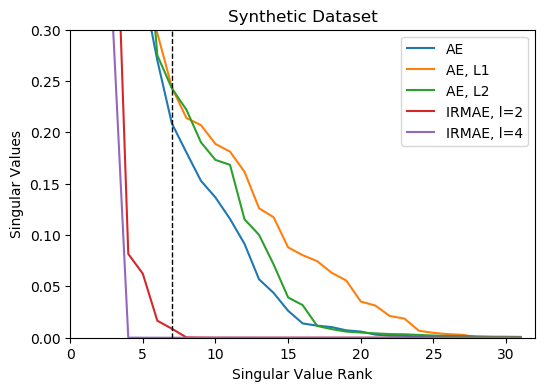}
    \caption{Singular values of the latent space of each model on synthetic shape dataset. Each curve represents singular values of the covariance matrix of the code computed on the validation set. IRMAE $l=2$ is able to approach the minimal theoretical rank of 7.}
    \label{fig:synthetic}
\end{figure}

The base architecture we used is a deterministic autoencoder. The architecture and experimental detail can be found in supplementary material. We use a latent dimension of 32. For IRMAE, we use $l=2$ and $l=4$ extra matrices between the encoder and the decoder. We test our method against non-regularization, L1 regularization, and L2 regularization on the hidden code with the same architecture. We demonstrate the learned latent space in Figure \ref{fig:synthetic}. The baseline model, L1 regularization, L2 regularization, IRMAE with $l=2$ yields excellent reconstructions on validation set.

This result shows that IRMAE with $l=2$ is able to learn good latent representation with a rank close to intrinsic dimension, while L1, L2 regularization tends to use a much larger latent space. 

\begin{figure}[h!]
\centering
\begin{tabular}{rc}
Baseline &
\includegraphics[width=0.7\textwidth]{ 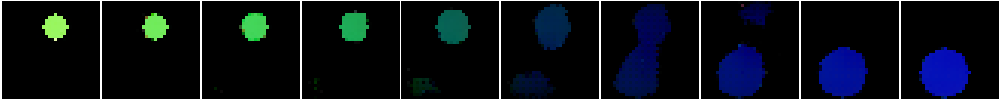}\\
L1 &
\includegraphics[width=0.7\textwidth]{ 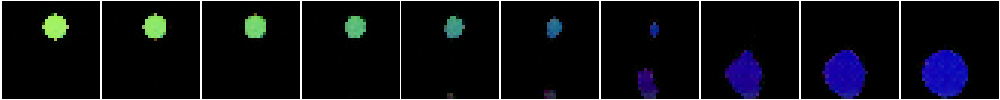}\\
L2 &
\includegraphics[width=0.7\textwidth]{ 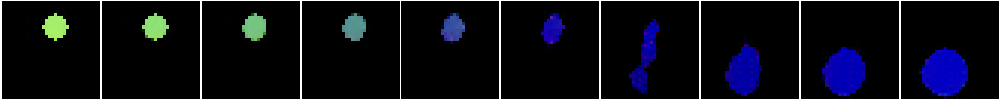}\\
IRMAE l=2 &
\includegraphics[width=0.7\textwidth]{ 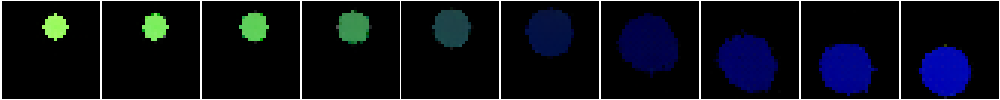}\\
IRMAE l=4 &
\includegraphics[width=0.7\textwidth]{ 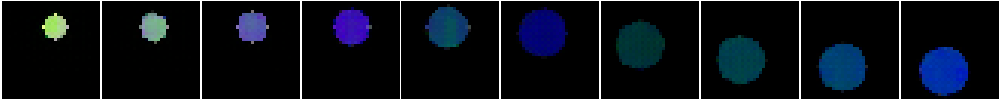}\\
\end{tabular}
\caption{Linear interpolation between two randomly generated samples. From top to bottom are results from baseline unregularized AE, AE with L1 regularization, AE with L2 regularization, IRMAE $l=2$, IRMAE $l=4$.}
\label{fig:synthetic-interpolation}
\end{figure}

\subsection{Image Generation}

Generating high-quality images by sampling the latent space is one of the key indicators of a good representation. In order to provide a comparison with standard deterministic autoencoders and variational autoencoders \cite{Kingma2014AutoEncodingVB}, we train our model on the MNIST dataset \cite{LeCun1998GradientbasedLA} and the CelebA dataset \cite{Liu2015DeepLF}. We set the latent dimension to 128/512 for the two datasets, respectively. We use 8/4 extra linear matrices for regularization in IRMAE, respectively. More experiment detail can be found in the supplementary material. We evaluate our model on a variety of representation learning tasks: interpolation between data points, sample generation from random noise, downstream classification task, PCA interpolation in latent space. We also quantitatively evaluate the sample generation by using the FID score. Each model uses the same architecture, except that the VAE code is twice as large to include the means and variances. On all these tasks, our method demonstrates comparable performances to the VAE. 

\begin{figure}[h!]
    \centering
    \includegraphics[height=1.8in]{ 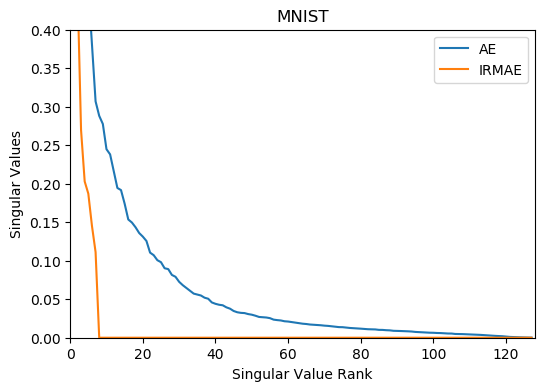}
    \includegraphics[height=1.8in]{ 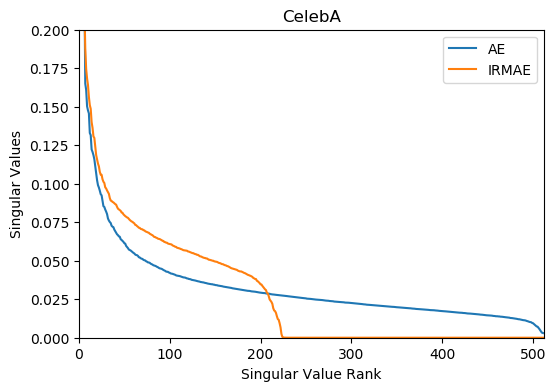}
    \caption{Singular value spectra of covariance matrices of codes for MNIST and CelebA datasets by IRMAE and a baseline AE. Each curve represents the singular values of the covariance matrix of the hidden code computed on the validation set.}
    \label{fig:sv}
\end{figure}

{\bf Latent Dimension}
We show the latent dimensionality reduction of our method in Figure \ref{fig:sv}. IRMAE utilizes significantly lower-dimensional latent space compared to baseline autoencoder. Notice that we omit the VAE's curve because VAE uses the whole latent space and hence all singular values tend to be large.

\begin{figure}[h!]
\centering
\settoheight{\tempdima}{\includegraphics[width=.24\linewidth]{example-image-a}}
\begin{tabular}{@{}c@{ }c@{ }c@{ }}
& MNIST \\ 
\rowname{AE}&
\includegraphics[width=0.6\linewidth]{ 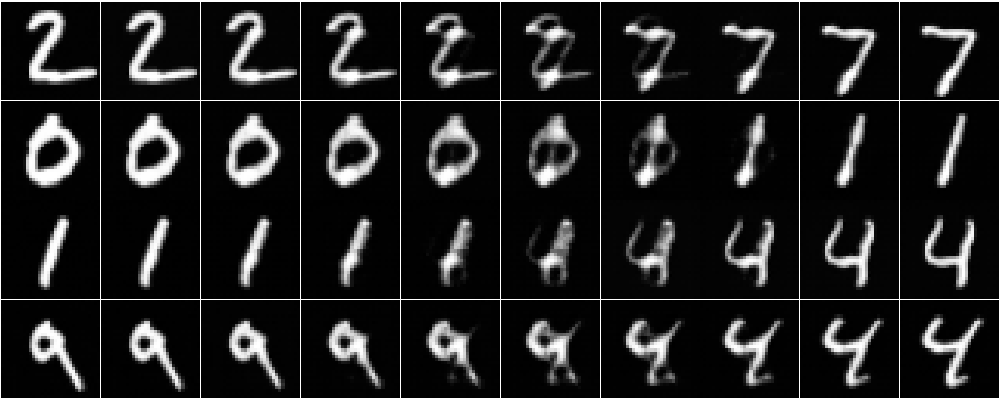}\\
\rowname{VAE}&
\includegraphics[width=0.6\linewidth]{ 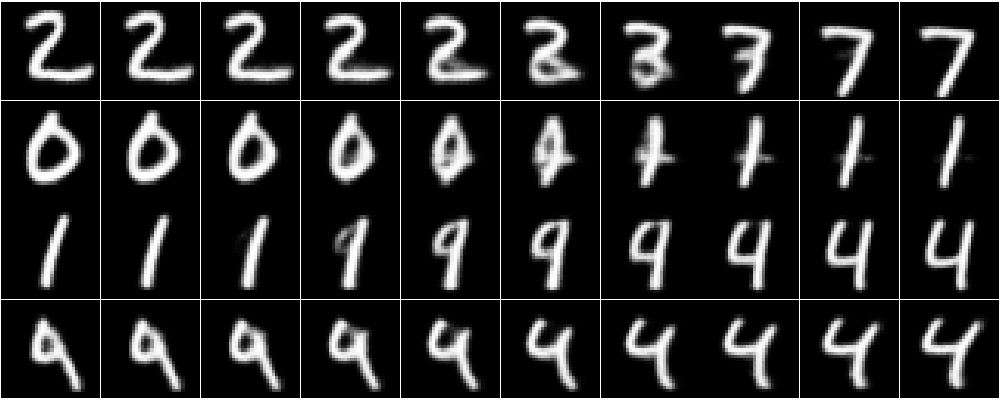}\\
\rowname{IRMAE}&
\includegraphics[width=0.6\linewidth]{ 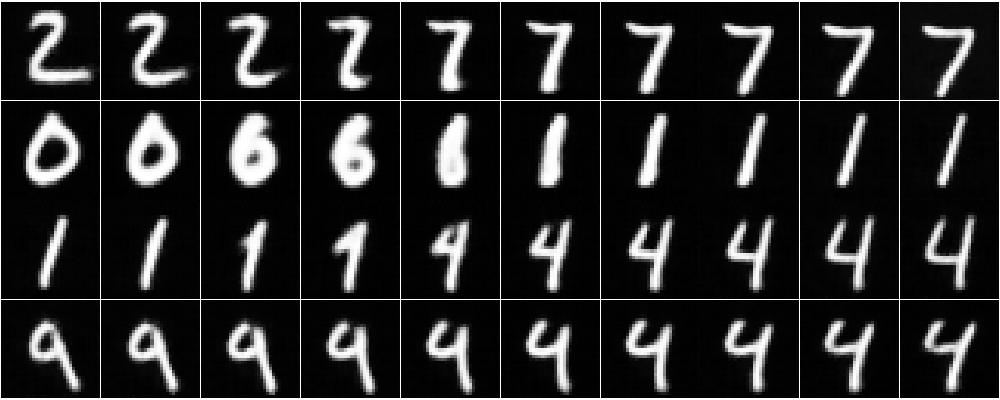}\\
\end{tabular}
\caption{Linear interpolation between data points on the MNIST dataset. From top to bottom are images generated from an unregularized AE, a VAE, and an IRMAE, respectively. IRMAE produces higher quality images.}
\label{fig:interpolate}
\end{figure}

{\bf Interpolation between Data Points:}
We linearly interpolate the latent variable between two images from the validation set. The generated results are shown in Figure \ref{fig:interpolate}. IRMAE significantly outperforms the baseline AE on MNIST. 

\begin{figure}[h!]
\centering
\settoheight{\tempdima}{\includegraphics[width=.4\linewidth]{example-image-a}}
\begin{tabular}{@{}c@{ }c@{ }c@{ }c@{ }}
& AE  & VAE & IRMAE \\  
\rowname{MNIST}&
\includegraphics[width=0.3\textwidth]{ 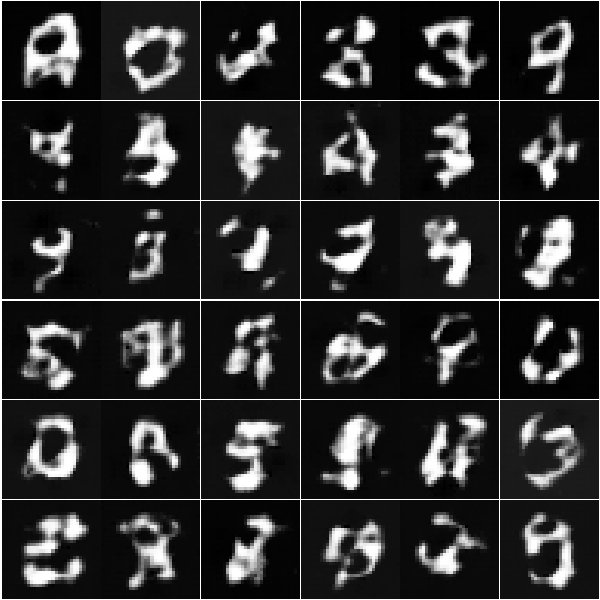} &
\includegraphics[width=0.3\textwidth]{ 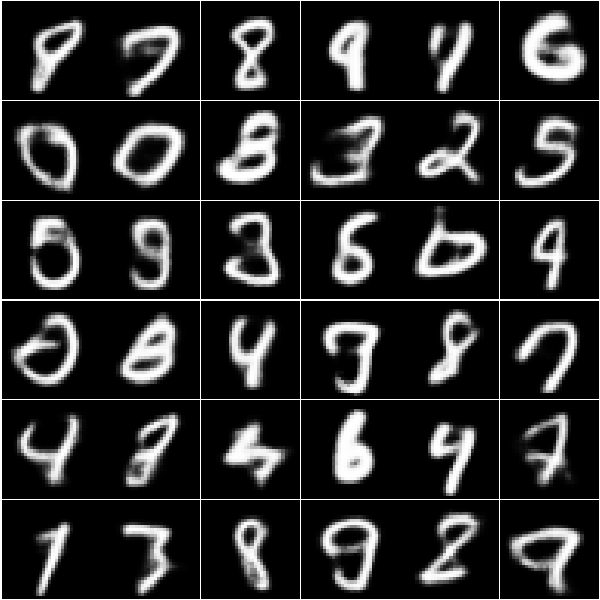} &
\includegraphics[width=0.3\textwidth]{ 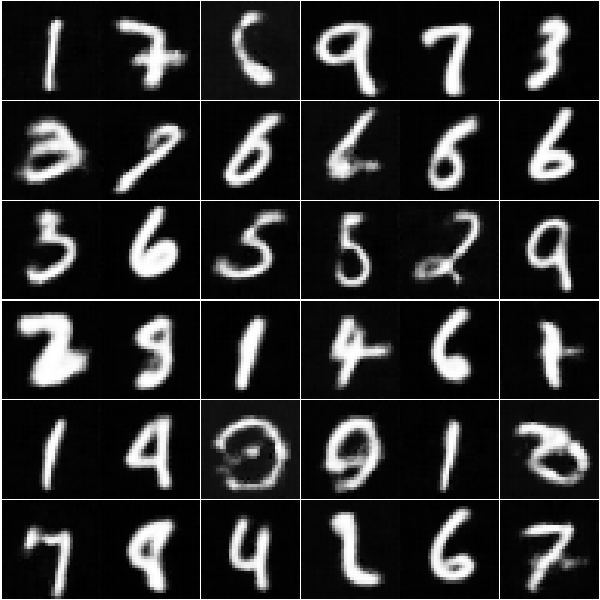}\\
\rowname{CelebA}&
\includegraphics[width=0.3\textwidth]{ 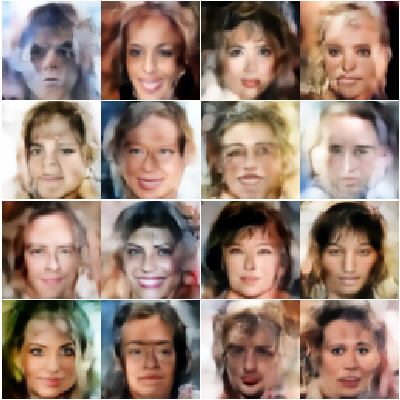} &
\includegraphics[width=0.3\textwidth]{ 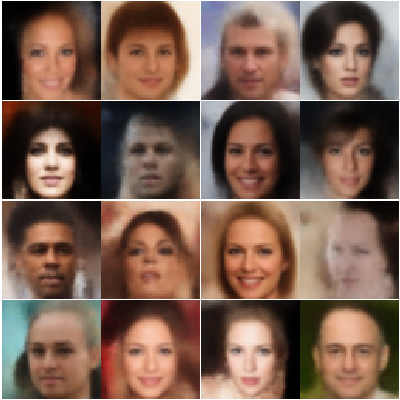} &
\includegraphics[width=0.3\textwidth]{ 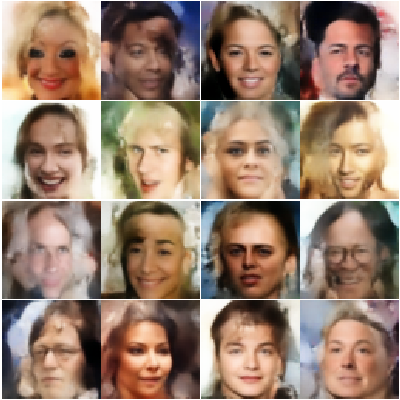}\\
\end{tabular}
\caption{MNIST/CelebA images samples from Multivariate Gaussian with covariance estimated from training set. From left to right are images generated from an unregularized AE, a VAE, and an IRMAE, respectively.}
\label{fig:mvg}
\end{figure}

\begin{figure}[h!]
\centering
\settoheight{\tempdima}{\includegraphics[width=.4\linewidth]{example-image-a}}
\begin{tabular}{@{}c@{ }c@{ }c@{ }c@{ }}
& AE  & VAE & IRMAE \\  
\rowname{MNIST}&
\includegraphics[width=0.3\textwidth]{ 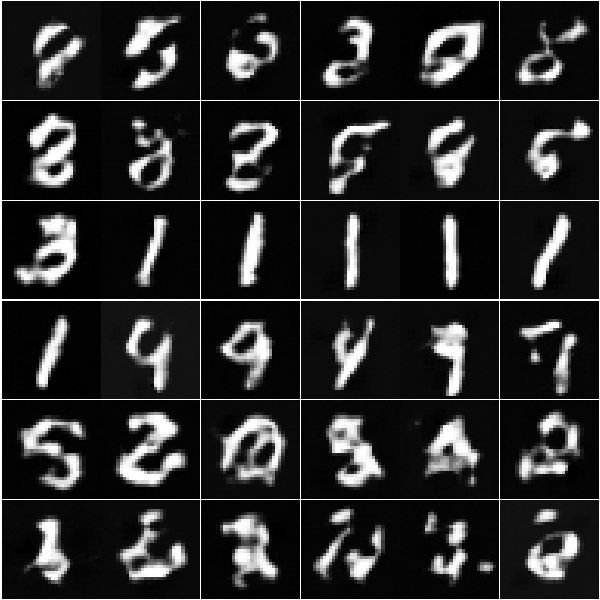} &
\includegraphics[width=0.3\textwidth]{ 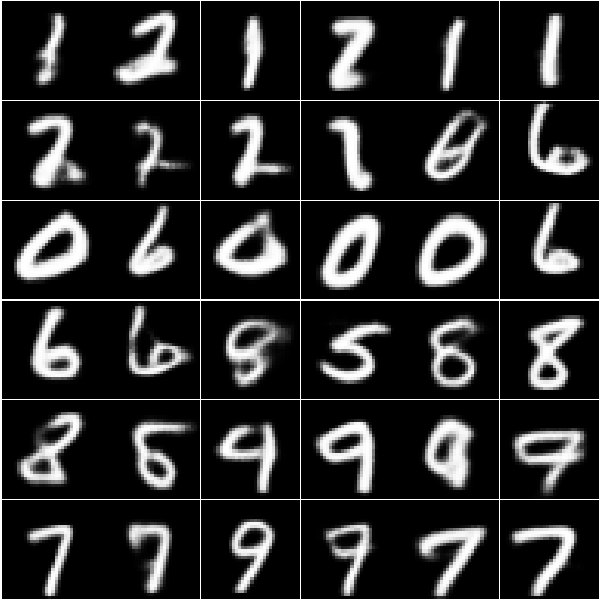} &
\includegraphics[width=0.3\textwidth]{ 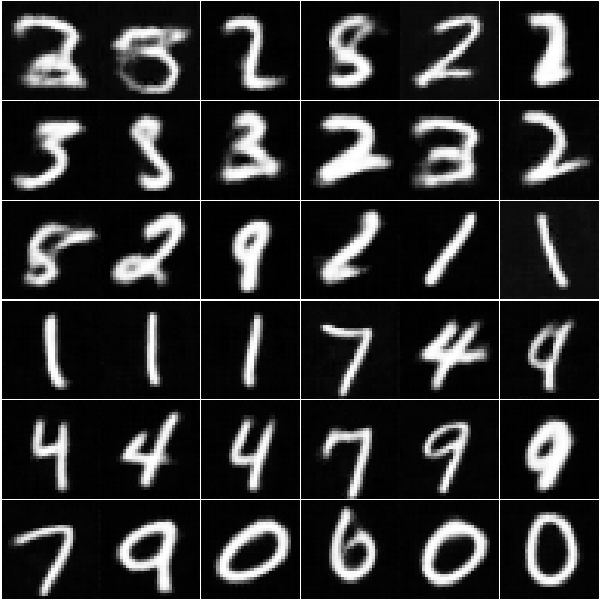}\\
\rowname{CelebA}&
\includegraphics[width=0.3\textwidth]{ 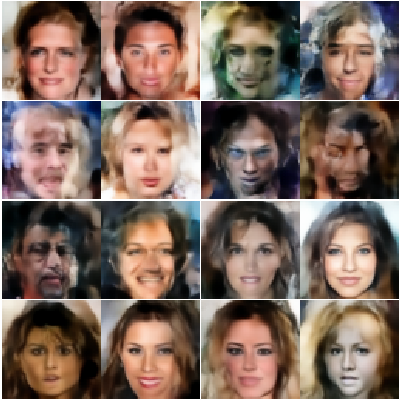} &
\includegraphics[width=0.3\textwidth]{ 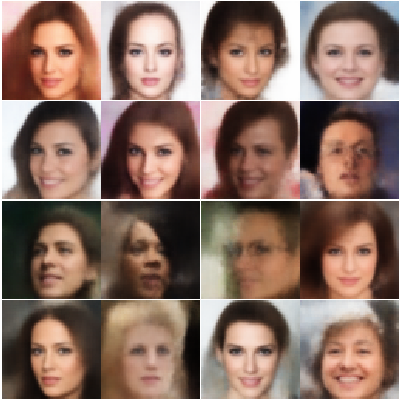} &
\includegraphics[width=0.3\textwidth]{ 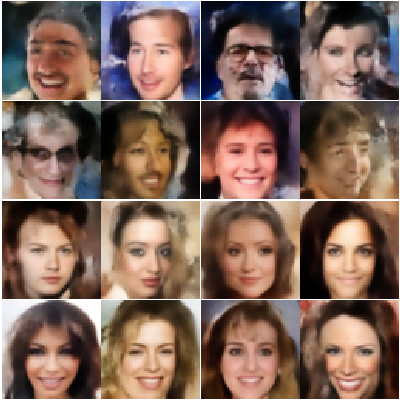}\\
\end{tabular}
\caption{MNIST/CelebA images samples from Gaussian Mixture Model with 4/10 clusters. From left to right are images generated from an unregularized AE, a VAE, and an IRMAE, respectively.}
\label{fig:gmm}
\end{figure}

{\bf Sampling from Noise:}
Deterministic autoencoders are not considered to be generative models. It is essential to have constraints on the latent space to derive such ability \cite{Bengio2013RepresentationLA}. Here, we show that IRMAE can sample high-quality images from Gaussian noise. Specifically, we sample random latent variables from 1) a multivariate Gaussian captured by this covariance matrix, 2) a Gaussian Mixture Model with 4/10 clusters. The generated results are presented in Figure \ref{fig:mvg} and Figure \ref{fig:gmm}. We quantitatively evaluate the performance of each model by using the Frechet Inception Distance (FID) \cite{Heusel2017GANsTB} and report the results on MNIST/CelebA in Table \ref{tbl:fid}.

\begin{table}[h!]
\caption{FID score (smaller is better) for samples of various models for MNIST/CelebA.}
\centering
\begin{tabular}{cc}
Multivariate Gaussian & Gaussian Mixture Model\\
\begin{tabular}{l|c|c|c}
\toprule
 & AE & VAE & IRMAE\\
\midrule
MNIST & 55.0 & 33.9 & 37.4\\
CelebA & 52.8 & 51.8 & 42.5\\
\bottomrule
\end{tabular}
&
\begin{tabular}{l|c|c|c}
\toprule
 & AE & VAE & IRMAE\\
\midrule
MNIST & 38.0 & 30.8 & 34.0\\
CelebA & 49.0 & 48.8  & 36.4 \\
\bottomrule
\end{tabular}
\end{tabular}
\label{tbl:fid}
\end{table}

\begin{figure}[h!]
    \centering
\settoheight{\tempdima}{\includegraphics[width=.63\linewidth]{example-image-a}}
\begin{tabular}{c@{ }c@{ }c@{ }}
AE & VAE & IRMAE \\
\includegraphics[width=0.32\textwidth]{ 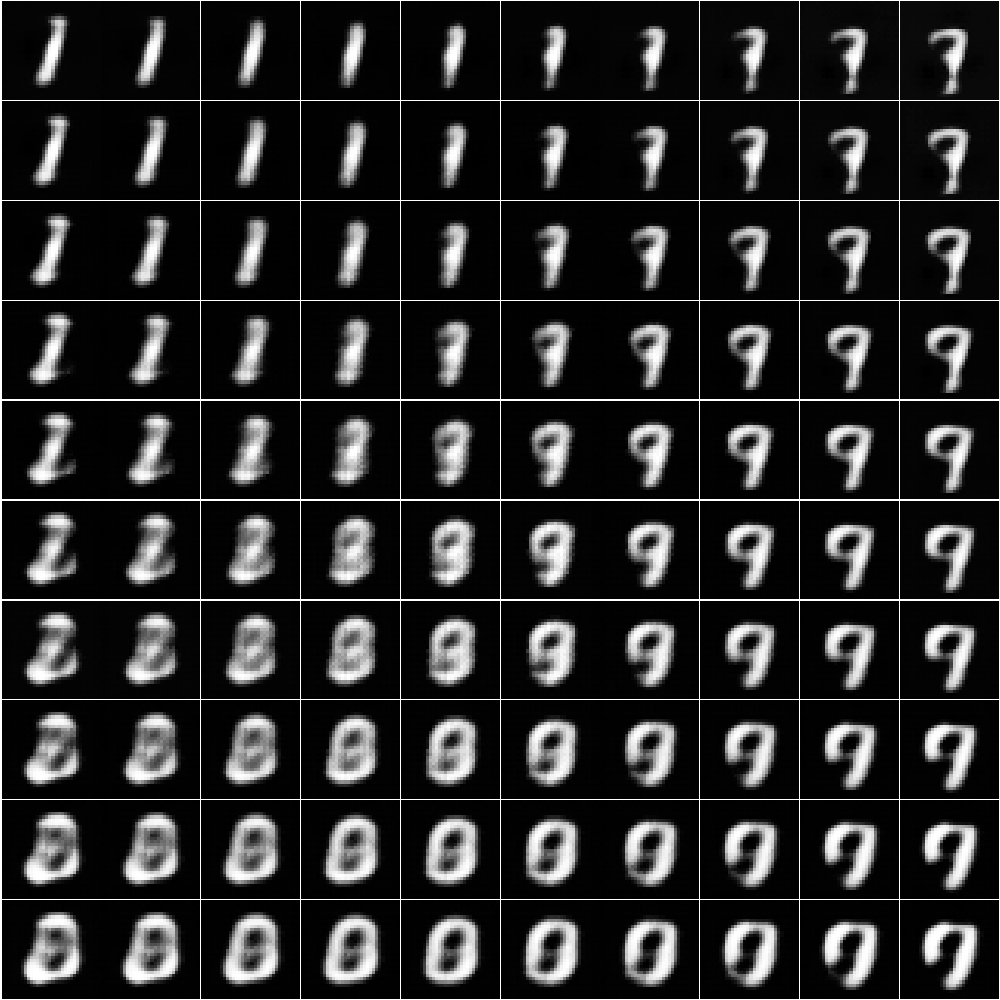} &
\includegraphics[width=0.32\textwidth]{ 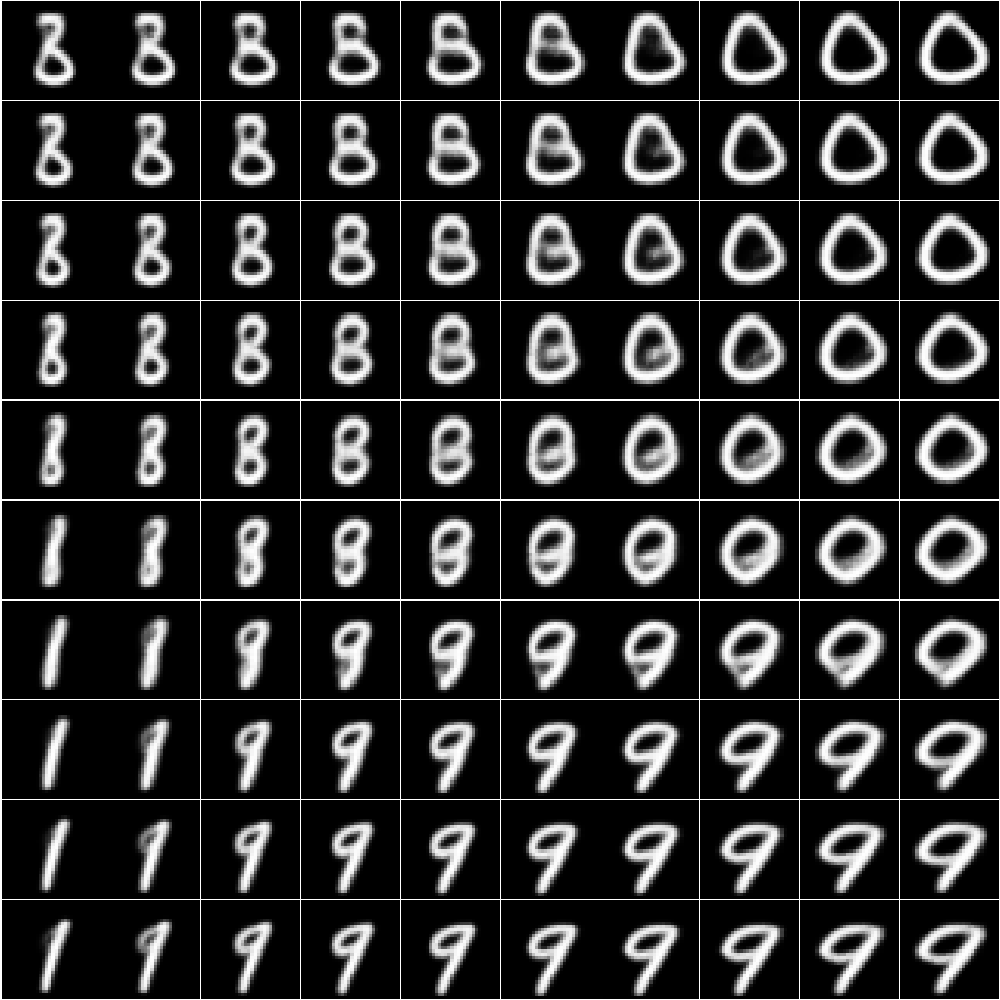} &
\includegraphics[width=0.32\textwidth]{ 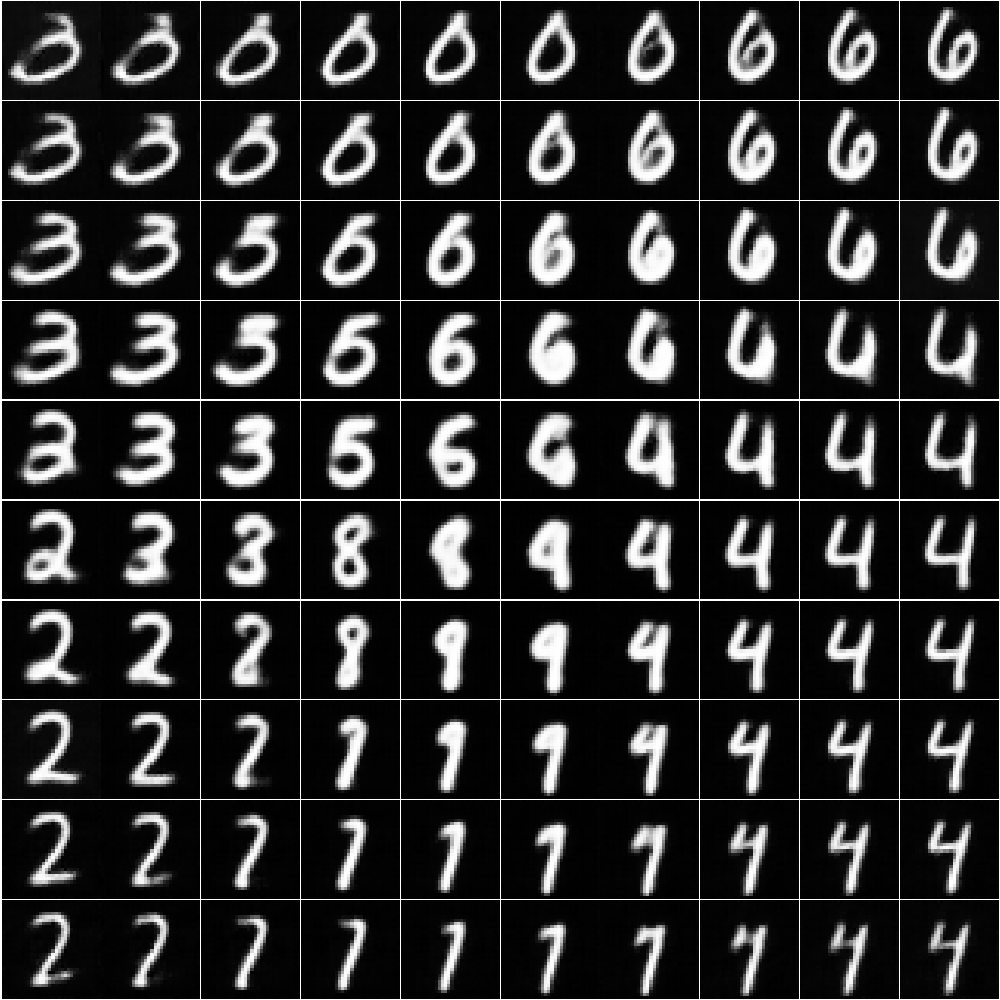}\\
\end{tabular}
\caption{Sampling images from 2-dimensional space, mapped by PCA from latent variables. We interpolate along two principal components to generate samples. From left to right are images generated from an unregularized AE, a VAE, and an IRMAE, respectively.}
    \label{fig:pca}
\end{figure}

{\bf PCA on Latent Space:}
We verify that IRMAE learns a compact and continuous latent space by performing PCA on the latent space. We project all latent variables to a 2-dimensional space. We randomly sample vectors in this low dimensional space and interpolate them along two principal vectors. The corresponding images are sampled from inverse PCA followed by the decoder, which is shown in Figure \ref{fig:pca}. IRMAE generates higher quality images compared to VAE.

Additional experiments are demonstrated in the supplementary material, including comparing IRMAE to other deterministic AEs, comparing IRMAE against AEs with various latent dimension, effect of varying linear layer depth in IRMAE.

\subsection{Downstream classification} 

Latent variables are useful for downstream tasks since they capture the main underlying structure of the data distribution \cite{He2019MomentumCF, Misra2019SelfSupervisedLO, Chen2020ASF}. These self-supervised learning methods have the exciting potential to outperform purely-supervised models. We train a multilayer perceptron head on the latent variable generated by the encoder, to classify MNIST images. This MLP head has two linear layers of hidden dimension 128, with ReLU activation. Thus, all models share the same architecture. Each model is trained with an Adam optimizer with a learning rate of 0.001. Early stopping is performed based on validation set accuracy. The encoder weights are kept fixed. We compare our method against several baselines as well as the supervised version whose entire network is trained jointly. Representations learned by IRMAE obtain a significantly lower error rate compared to those from the unregularized AE in this task. The results are listed in Table \ref{tbl:classificatin}. 

\begin{table}[h!]
\caption{Downstream classification on MNIST dataset. We add a MLP head on top of the pretrained encoder by each method. Thus, all models share the same architecture. We do not perform fine tuning on the pretrained encoder except with the purely supervised version. Representation learned by IRMAE obtains significantly lower error rate compared to baselines and supervised version in the low labeled data regime.}
\centering
\begin{tabular}{l|c|c|c|c|c}
\toprule
total training size & 10 & 100 & 1000 & 10000 & 60000 \\
\midrule
AE & 31.4$\pm$0.5 & 30.2$\pm$0.3 & 10.6$\pm$0.2 & 3.7$\pm$0.1 & 1.9$\pm$0.1 \\
VAE & 21.8$\pm$1.0 & 21.7$\pm$0.4 & 5.1$\pm$0.2 & {\bf 1.7$\pm$0.1} & 1.1$\pm$0.1 \\
IRMAE & {\bf 12.0$\pm$0.9} & {\bf 10.2$\pm$0.5} & {\bf 3.8$\pm$0.3} & 2.4$\pm$0.2 & 1.9$\pm$0.1 \\
\midrule
supervised & 29.1$\pm$2.6 & 25.1$\pm$0.6 & 6.0$\pm$0.4 & {\bf 1.7$\pm$0.1} & {\bf 0.8$\pm$0.1} \\
\bottomrule
\end{tabular}
\label{tbl:classificatin}
\end{table}

\subsection{Ablation Study}

We perform several ablation studies to verify that the effect of dimensionality reduction comes from the extra linear neural network and its optimization dynamics.

\begin{figure}[h!]
    \centering
    \includegraphics[width=0.5\textwidth]{ 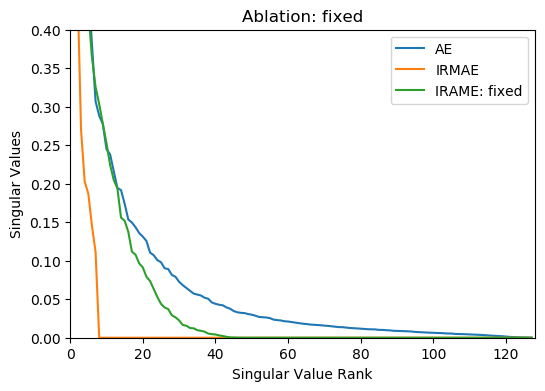}
    \includegraphics[width=0.35\textwidth]{ 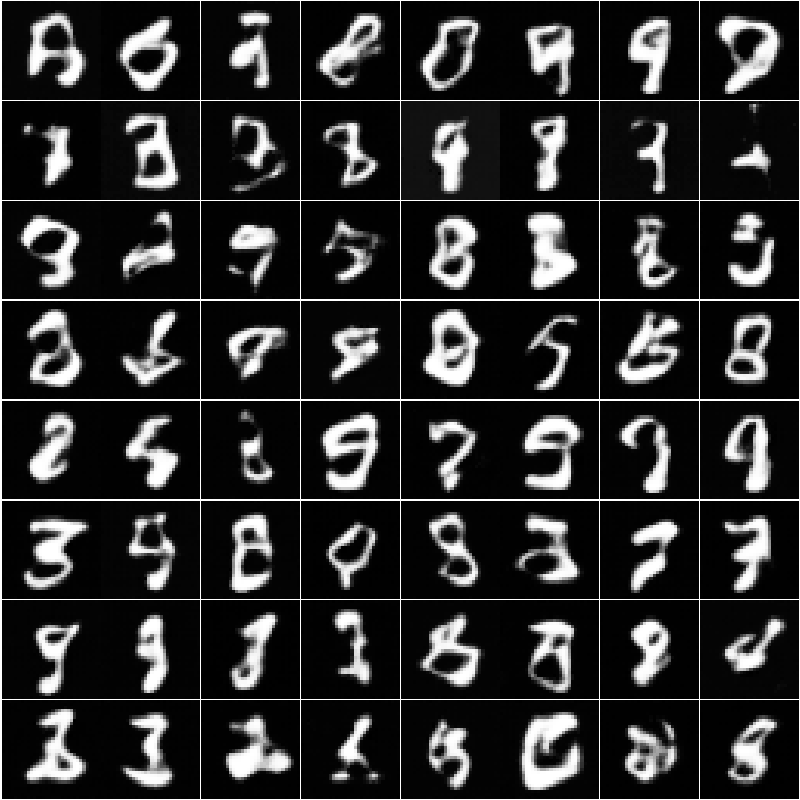}
    \caption{Ablation study: linear matrices fixed. This proves that the regularization behavior is not an effect of naive soft bottleneck.}
    \label{fig:ablation-fix}
\end{figure}

{\bf Linear matrices fixed:} In this ablation experiment, we fix the linear matrices to verify that the regularization effect comes from the learning dynamics instead of just the architecture. Figure \ref{fig:ablation-fix} shows that under this condition, the regularization effect is weakened, and the sampled images are significantly worse.

\begin{figure}[h!]
    \centering
    \includegraphics[width=0.5\textwidth]{ 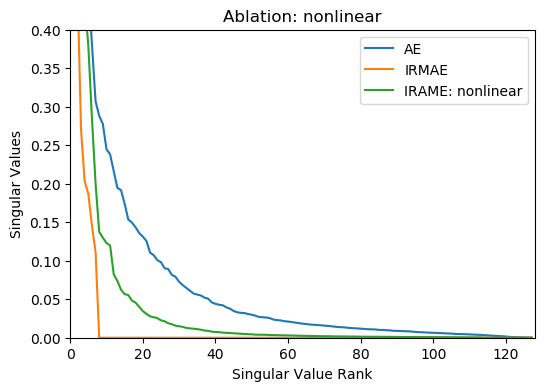}
    \includegraphics[width=0.35\textwidth]{ 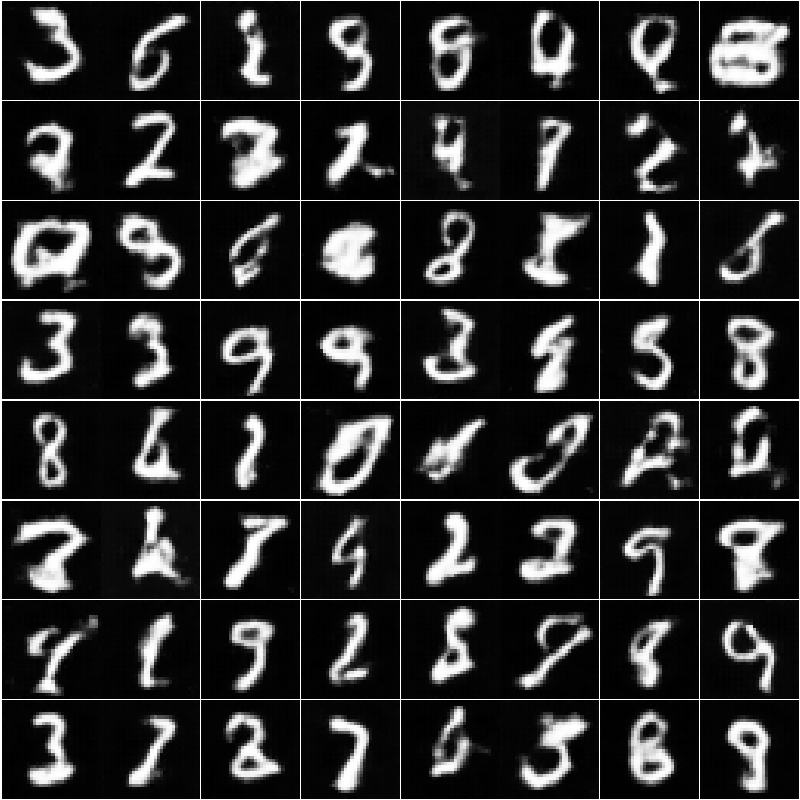}
    \caption{Ablation study: adding nonlinearity layers between linear matrices. This proves that the regularization behavior is not a naive effect of deeper architecture.}
    \label{fig:ablation-nonlinear}
\end{figure}

{\bf Nonlinearity between matrices:} One may suspect that the regularization effect comes from a deeper architecture. If we add nonlinearity between matrices, the model is equivalent to a standard autoencoder, with more layers. We show that adding a nonlinearity results in worse generation results, and the regularization effect is also completely lost. See Figure \ref{fig:ablation-nonlinear}.

\begin{figure}[h!]
    \centering
    \includegraphics[width=0.5\textwidth]{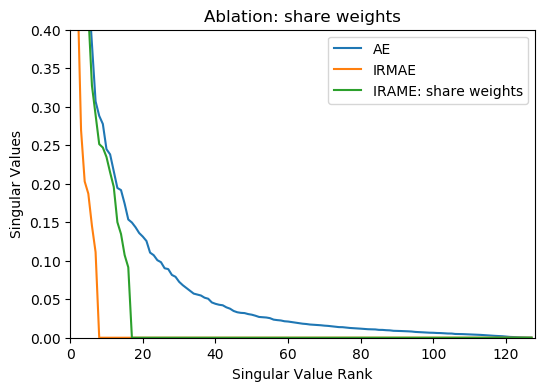}
    \includegraphics[width=0.35\textwidth]{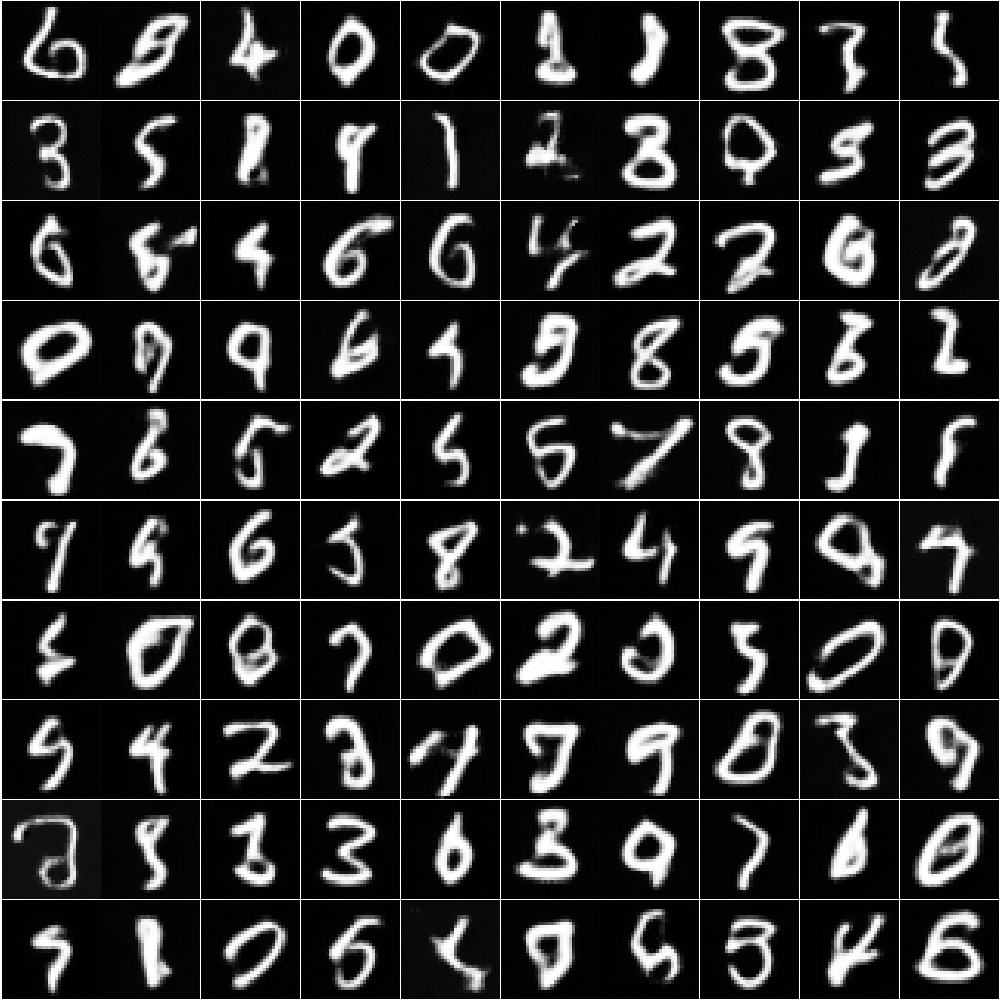}
    \caption{Ablation study: sharing weights in the inserted linear layers.}
    \label{fig:ablation-share}
\end{figure}

{\bf Weight Sharing:} As our method introduces more parameters for training, it would be desirable to have all inserted matrices to share weights to reduce memory requirement. We show that forcing all matrices to share weights results in slightly worse generation results and weakened regularization effect. See Figure 
\ref{fig:ablation-share}.

\section{Conclusion}

An important component of autoencoder methods is the method by which the information capacity of the latent representation is minimized or limited. In this work, the rank of the covariance matrix of the codes is implicitly minimized by relying on the fact that gradient descent learning in multi-layer linear networks leads to minimum-rank solutions. By inserting a number of extra linear layers between the encoder and the decoder, the system spontaneously learns representations with a low effective dimension. The model, dubbed Implicit Rank-Minimizing Autoencoder (IRMAE), is simple, deterministic, and low-rank latent space. We demonstrate the validity of the method on several image generation and representation learning tasks.

\section*{Broader Impact}

This work provides a novel approach to representation learning and self-supervised learning. It has the potential of boosting general self-supervised learning performances with social benefits including requiring less human data labeling, reducing power consumption of AI models, improving data privacy. 

\section*{Acknowledgement}
We are grateful to Stephane Deny for his feedback on early versions of the manuscript. We thank Pascal Vincent, Nicolas Ballas, Lluis Castrejon, Piotr Bojanowski for their
fruitful discussions.

\bibliography{citation}

\begin{thebibliography}{10}

\bibitem{Arora2019ImplicitRI}
Sanjeev Arora, Nadav Cohen, Wei Hu, and Yuping Luo.
\newblock Implicit regularization in deep matrix factorization.
\newblock In {\em Advances in Neural Information Processing Systems (NeurIPS
  '19)}, pages 7413--7424, 2019.

\bibitem{Bengio2013RepresentationLA}
Yoshua Bengio, Aaron~C. Courville, and Pascal Vincent.
\newblock Representation learning: A review and new perspectives.
\newblock {\em IEEE Transactions on Pattern Analysis and Machine Intelligence},
  35:1798--1828, 2013.

\bibitem{Chen2020ASF}
Ting Chen, Simon Kornblith, Mohammad Norouzi, and Geoffrey Hinton.
\newblock A simple framework for contrastive learning of visual
  representations.
\newblock {\em arXiv preprint arXiv:2002.05709}, 2020.

\bibitem{doi-nips-2004}
Eizaburo Doi and Michael~S. Lewicki.
\newblock Sparse coding of natural images using an overcomplete set of limited
  capacity units.
\newblock In L.~K. Saul, Y.~Weiss, and L.~Bottou, editors, {\em Advances in
  Neural Information Processing Systems 17}, pages 377--384. MIT Press, 2005.

\bibitem{Ghosh2020FromVT}
Partha Ghosh, Mehdi S.~M. Sajjadi, Antonio Vergari, Michael Black, and Bernhard
  Scholkopf.
\newblock From variational to deterministic autoencoders.
\newblock In {\em International Conference on Learning Representations (ICLR
  '20)}, 2020.

\bibitem{Gidel2019ImplicitRO}
Gauthier Gidel, Francis Bach, and Simon Lacoste-Julien.
\newblock Implicit regularization of discrete gradient dynamics in linear
  neural networks.
\newblock In {\em Advances in Neural Information Processing Systems 32 (NeurIPS
  '19)}, pages 3202--3211, 2019.

\bibitem{goroshin-lecun-iclr-13}
Rotislav Goroshin and Yann LeCun.
\newblock Saturating auto-encoders.
\newblock In {\em International Conference on Learning Representations
  (ICLR2013)}, April 2013.

\bibitem{Gunasekar2018ImplicitBO}
Suriya Gunasekar, Jason~D. Lee, Daniel Soudry, and Nathan Srebro.
\newblock Implicit bias of gradient descent on linear convolutional networks.
\newblock In {\em Advances in Neural Information Processing Systems (NeurIPS
  '18)}, pages 9461--9471, 2018.

\bibitem{Gunasekar2018ImplicitRI}
Suriya Gunasekar, Blake~E Woodworth, Srinadh Bhojanapalli, Behnam Neyshabur,
  and Nati Srebro.
\newblock Implicit regularization in matrix factorization.
\newblock In {\em Advances in Neural Information Processing Systems 30 (NeurIPS
  '17)}, pages 6151--6159, 2017.

\bibitem{He2019MomentumCF}
Kaiming He, Haoqi Fan, Yuxin Wu, Saining Xie, and Ross Girshick.
\newblock Momentum contrast for unsupervised visual representation learning.
\newblock {\em arXiv preprint arXiv:1911.05722}, 2019.

\bibitem{Heusel2017GANsTB}
Martin Heusel, Hubert Ramsauer, Thomas Unterthiner, Bernhard Nessler, and Sepp
  Hochreiter.
\newblock Gans trained by a two time-scale update rule converge to a local nash
  equilibrium.
\newblock In {\em Advances in Neural Information Processing Systems 30 (NeurIPS
  '17)}, pages 6626--6637, 2017.

\bibitem{Higgins2017betaVAELB}
Irina Higgins, Lo{\"i}c Matthey, Arka Pal, Christopher Burgess, Xavier Glorot,
  Matthew~M Botvinick, Shakir Mohamed, and Alexander Lerchner.
\newblock beta-vae: Learning basic visual concepts with a constrained
  variational framework.
\newblock In {\em International Conference on Learning Representations (ICLR
  '20)}, 2020.

\bibitem{Kingma2015AdamAM}
Diederik~P. Kingma and Jimmy Ba.
\newblock Adam: A method for stochastic optimization.
\newblock {\em CoRR}, abs/1412.6980, 2015.

\bibitem{Kingma2014AutoEncodingVB}
Diederik~P. Kingma and Max Welling.
\newblock Auto-encoding variational bayes.
\newblock {\em arXiv preprint arXiv:1312.6114}, 2014.

\bibitem{LeCun1998GradientbasedLA}
Yann LeCun, Léon Bottou, Yoshua Bengio, and Patrick Haffner.
\newblock Gradient-based learning applied to document recognition.
\newblock In {\em Proceedings of the IEEE}, volume~86, pages 2278--2324, 1998.

\bibitem{Liu2015DeepLF}
Ziwei Liu, Ping Luo, Xiaogang Wang, and Xiaoou Tang.
\newblock Deep learning face attributes in the wild.
\newblock {\em 2015 IEEE International Conference on Computer Vision (ICCV
  '15)}, pages 3730--3738, 2015.

\bibitem{Maaten2008VisualizingDU}
L.~V.~D. Maaten and Geoffrey~E. Hinton.
\newblock Visualizing data using t-sne.
\newblock {\em Journal of Machine Learning Research}, 9:2579--2605, 2008.

\bibitem{Misra2019SelfSupervisedLO}
Ishan Misra and Laurens van~der Maaten.
\newblock Self-supervised learning of pretext-invariant representations.
\newblock In {\em Conference on Computer Vision and Pattern Recognition (CVPR
  '20)}, 2020.

\bibitem{Ng2000SparseAE}
Andrew Ng.
\newblock Sparse autoencoder.
\newblock 2000.

\bibitem{ranzato-nips-07}
Marc'Aurelio Ranzato, {Y-Lan} Boureau, and Yann LeCun.
\newblock Sparse feature learning for deep belief networks.
\newblock In {\em Advances in Neural Information Processing Systems (NIPS
  2007)}, volume~20, 2007.

\bibitem{Razin2020ImplicitRI}
Noam Razin and Nadav Cohen.
\newblock Implicit regularization in deep learning may not be explainable by
  norms.
\newblock {\em arXiv preprint arXiv:2005.06398}, 2020.

\bibitem{Rifai2011ContractiveAE}
Salah Rifai, Pascal Vincent, Xavier Muller, Xavier Glorot, and Yoshua Bengio.
\newblock Contractive auto-encoders: Explicit invariance during feature
  extraction.
\newblock In {\em International Conference on Machine Learning (ICML '11)},
  2011.

\bibitem{rhw-1986}
D.~E. Rumelhart, G.~E. Hinton, and R.~J. Williams.
\newblock Learning internal representations by error propagation.
\newblock In {\em Parallel Distributed Processing: Explorations in the
  Microstructure of Cognition}, pages 318--362. MIT Press, 1986.

\bibitem{Saxe2019AMT}
Andrew~M. Saxe, James~L. McClelland, and Surya Ganguli.
\newblock A mathematical theory of semantic development in deep neural
  networks.
\newblock {\em Proceedings of the National Academy of Sciences of the United
  States of America}, 116 23:11537--11546, 2019.

\bibitem{Soudry2018TheIB}
Daniel Soudry, Elad Hoffer, and Nathan Srebro.
\newblock The implicit bias of gradient descent on separable data.
\newblock In {\em International Conference on Learning Representations (ICLR
  '18)}, 2018.

\bibitem{Tolstikhin2018WassersteinA}
Ilya Tolstikhin, Olivier Bousquet, Sylvain Gelly, and Bernhard Schoelkopf.
\newblock Wasserstein auto-encoders.
\newblock In {\em International Conference on Learning Representations (ICLR
  '18)}, 2018.

\bibitem{Oord2017NeuralDR}
A{\"a}ron van~den Oord, Oriol Vinyals, and Koray Kavukcuoglu.
\newblock Neural discrete representation learning.
\newblock In {\em Advances in Neural Information Processing Systems (NeurIPS
  '17)}, 2017.

\bibitem{Vincent2008ExtractingAC}
Pascal Vincent, Hugo Larochelle, Yoshua Bengio, and Pierre-Antoine Manzagol.
\newblock Extracting and composing robust features with denoising autoencoders.
\newblock In {\em International Conference on Machine Learning (ICML '08)},
  2008.

\end{thebibliography}
\bibliographystyle{plain}

\newpage
\section*{Appendix}

\subsection{Experiment Detail}

\subsubsection{Dataset}
For the synthetic shape dataset, we generate shape images on the fly. The size of each shape is uniformly sampled between 3 and 8, inclusively. The color is uniformly sampled in RGB. The coordinate of the center of the shape is randomly sampled with x and y between 8 and 24, inclusively. 

For the MNIST dataset, all images are resized to 32x32.

For the CelebA dataset, all images are center-chopped to 148x148 and then resized to 64x64.

\subsubsection{Architecture}

The architecture of the encoder and the decoder for each experiment is listed in Tables \ref{tbl:architecture}. Conv$_n$/ConvT denotes a convolutional/transposed-convolutional layer with the output channel dimension equal to $n$. All convolutional layers use 4x4 kernel size with a stride 2, padding 1. FC$_n$ denotes a fully connected network with output dimension $n$.

\begin{table}[h!]
\caption{The architecture of the encoder and the decoder for each experiment.}
\centering
\begin{tabular}{l|c|c|c}
\toprule
Dataset & Shape & MNIST & CelebA\\
\midrule
Encoder & \begin{tabular}{l}
$x \in \mathcal{R}^{32x32x3}$ \\
$\rightarrow$ Conv$_{32}$
$\rightarrow$
ReLU  \\
$\rightarrow$ Conv$_{64}$
$\rightarrow$
ReLU  \\    $\rightarrow$ Conv$_{128}$
$\rightarrow$
ReLU  \\
$\rightarrow$ Conv$_{256}$
$\rightarrow$
ReLU  \\ 
$\rightarrow$ Conv$_{32}$
$\rightarrow$
ReLU  \\ 
$\rightarrow$
$z\in\mathcal{R}^{32}$
\end{tabular} & 
\begin{tabular}{l}
$x \in \mathcal{R}^{32x32x1}$ \\
$\rightarrow$ Conv$_{32}$
$\rightarrow$
ReLU  \\
$\rightarrow$ Conv$_{64}$
$\rightarrow$
ReLU  \\    $\rightarrow$ Conv$_{128}$
$\rightarrow$
ReLU  \\
$\rightarrow$ Conv$_{256}$
$\rightarrow$
ReLU  \\ 
$\rightarrow$
flattern$\_$to 1024\\
$\rightarrow$
FC$_{128}$
$\rightarrow$
$z\in\mathcal{R}^{128}$
\end{tabular}& 
\begin{tabular}{l}
$x \in \mathcal{R}^{64x64x3}$ \\
$\rightarrow$ Conv$_{128}$
$\rightarrow$
ReLU  \\
$\rightarrow$ Conv$_{256}$
$\rightarrow$
ReLU  \\    $\rightarrow$ Conv$_{512}$
$\rightarrow$
ReLU  \\
$\rightarrow$ Conv$_{1024}$
$\rightarrow$
ReLU  \\ 
$\rightarrow$
flattern$\_$to 16384\\ 
$\rightarrow$
FC$_{512}$$\rightarrow$
$z\in\mathcal{R}^{512}$
\end{tabular}\\
\midrule
Decoder & 
\begin{tabular}{l}
$z\in\mathcal{R}^{32}$\\
$\rightarrow$
ConvT$_{256}$
$\rightarrow$
ReLU\\
$\rightarrow$
ConvT$_{128}$
$\rightarrow$
ReLU\\
$\rightarrow$
ConvT$_{64}$
$\rightarrow$
ReLU\\
$\rightarrow$
ConvT$_{32}$
$\rightarrow$
ReLU\\
$\rightarrow$
ConvT$_{3}$
$\rightarrow$
Tanh\\
$\rightarrow$
$\hat{x}\in\mathcal{R}^{32x32x3}$
\end{tabular}
& 
\begin{tabular}{l}
$z\in\mathcal{R}^{128}$\\
$\rightarrow$
FC$_{8096}$\\
$\rightarrow$
reshape$\_$to 8x8x128\\
$\rightarrow$
ConvT$_{64}$
$\rightarrow$
ReLU\\
$\rightarrow$
ConvT$_{32}$
$\rightarrow$
ReLU\\
$\rightarrow$
ConvT$_{3}$
$\rightarrow$
Tanh\\
$\rightarrow$
$\hat{x}\in\mathcal{R}^{32x32x1}$
\end{tabular}
& 
\begin{tabular}{l}
$z\in\mathcal{R}^{512}$\\
$\rightarrow$
FC$_{65536}$\\
$\rightarrow$
reshape$\_$to 8x8x1024\\
$\rightarrow$
ConvT$_{512}$
$\rightarrow$
ReLU\\
$\rightarrow$
ConvT$_{256}$
$\rightarrow$
ReLU\\
$\rightarrow$
ConvT$_{128}$
$\rightarrow$
ReLU\\
$\rightarrow$
ConvT$_{3}$
$\rightarrow$
Tanh\\
$\rightarrow$
$\hat{x}\in\mathcal{R}^{64x64x3}$
\end{tabular}
\\
\bottomrule
\end{tabular}
\label{tbl:architecture}
\end{table}

For VAE models, the last layer of the decoder has doubled output dimension, which is split as the average and the standard deviation. It also uses Sigmoid instead of Tanh.

\subsubsection{Hyperparameters}

The following hyperparameters for each experiment are listed in Table. \ref{tbl:hyperparameter}. The number of epochs is chosen for converged reconstruction error for the base model.

\begin{table}[h!]
\caption{hyperparameters.}
\centering
\begin{tabular}{l|c|c|c}
\toprule
Dataset & Shape & MNIST & CelebA\\
\midrule
learning rate & 0.0001 & 0.0001 & 0.0001 \\
epochs & 100 & 50 & 100 \\
latent dimension & 32 & 128 & 512 \\
batch size & 32 & 32 & 32 \\
training examples & 50000 & 60000 & 162770 \\ 
evaluation examples & 10000 & 10000 & 19962\\
\bottomrule
\end{tabular}
\label{tbl:hyperparameter}
\end{table}

\subsection{Additional Experiments}

\subsubsection{Effect of Varying Linear Layers Initial Variance}
Initial variance of the linear matrices has strong influence on the regularization effect. We observe that a larger variance weakens the regularization effect. See Table.\ref{tbl:var}.

\begin{table}[h!]
    \centering
    \caption{Effect of varying initial variance of linear layers in IRMAE. Performed on MNIST dataset. Latent rank represents corresponding number of nonzero singular values of the covariance matrix of latent space.}
    \begin{tabular}{c|c|c|c}
    \toprule
    Variance & 1x & 2x & 4x \\
    \midrule
    Latent Rank & 8 & 43 & 66 \\ 
    \midrule
    FID & 37.4 & 33.8 & 49.0 \\
    \bottomrule
    \end{tabular}
\label{tbl:var}
\end{table}

\subsubsection{Effect of Varying Linear Layers Depth}

Adding more linear layers will increase the regularization effect. We demonstrate such effect in Table.\ref{tbl:depth}.
The number of linear layers $l$ is a hyperparameter and needs to be optimized in practice.

\begin{table}[h!]
\centering
\caption{Effect of varying linear layers depth. Performed on MNIST dataset. Latent rank represents corresponding number of nonzero singular values of the covariance matrix of latent space.}
\begin{tabular}{c|c|c|c|c}
\toprule
Depth (l)		&		2	&	4	&	8	&	12 \\
\midrule
Latent Rank	&		70	&	39	&	8	&	4 \\
\midrule
FID	&		44.0	&	30.1	&	37.4   &		62.6 \\
\bottomrule
\end{tabular}
\label{tbl:depth}
\end{table}

\subsubsection{Comparing to State-of-the-art Deterministic AEs}

We compare IRMAE against several modern deterministic autoencders including WAE and RAE. IRMAE demonstrates superior performance on CelebA dataset. See Table.\ref{tbl:sota}.

\begin{table}[h!]
\small
\centering
    \caption{Comparing IRMAE against state-of-the-art deterministic AEs on CelebA dataset.}
    \begin{tabular}{c|c|c}
    \toprule
    WAE \cite{Tolstikhin2018WassersteinA} & RAE \cite{Ghosh2020FromVT} & IRMAE \\
    \midrule
    53.7 & 44.7 & \textbf{42.0}\\
    \bottomrule
    \end{tabular}
\label{tbl:sota}
\end{table}

\subsubsection{Comparing to AEs with Various Latent Dimension}

\begin{figure}[h!]
    \centering
    \includegraphics[width=0.8\textwidth]{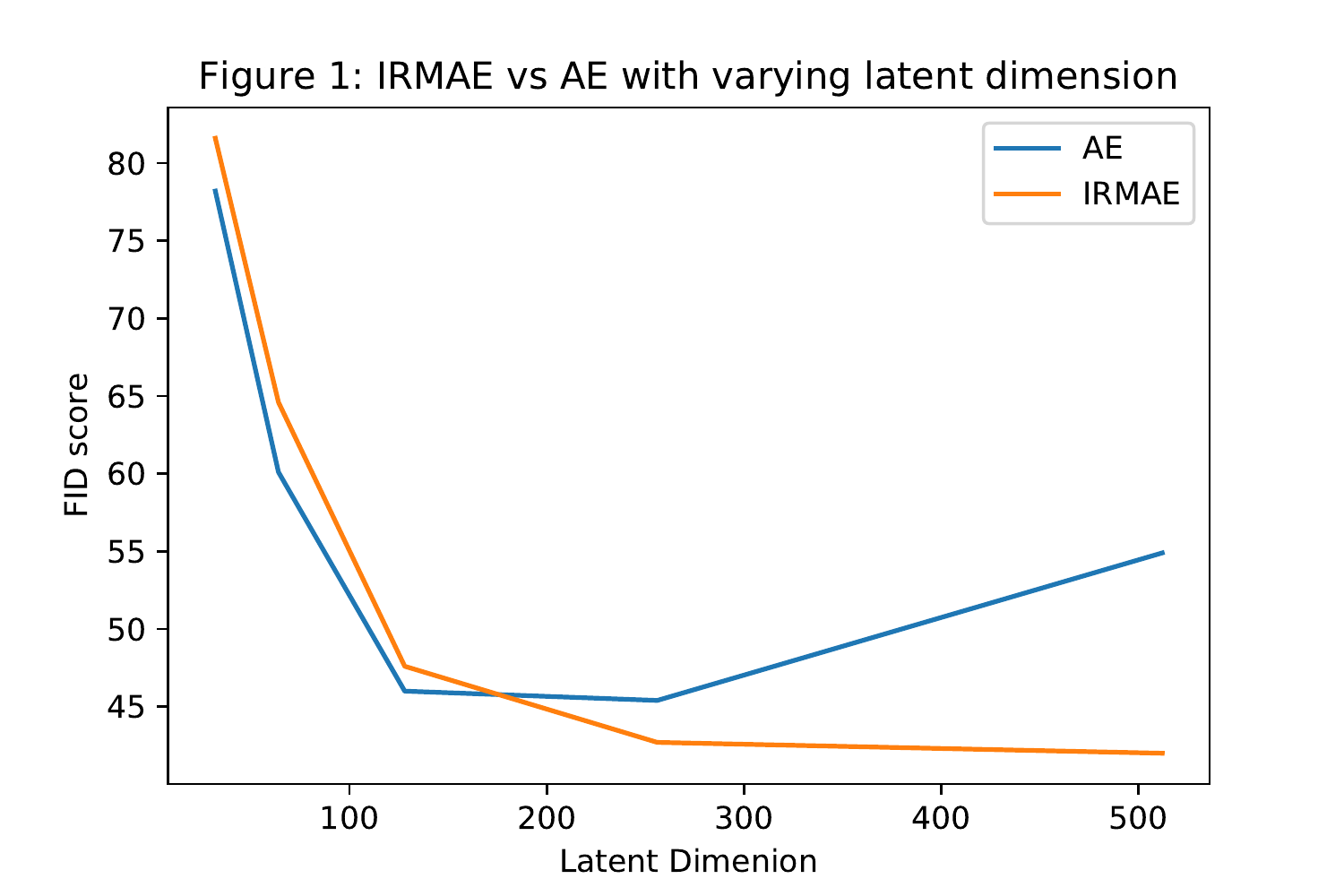}
    \caption{Comparing IRMAE against AEs with different latent dimension. Performed on CelebA dataset.}
    \label{fig:dim}
\end{figure}

Autoencoders with different latent dimension or prior setting has trade-off in learning useful representations. Here, we study the effect of latent dimensionality of IRMAE against AE in Table.\ref{tbl:dim} and Figure.\ref{fig:dim}. IRMAE with larger latent dimensions outperforms the optimal dimensional AE.

\begin{table}[h!]
    \centering
    \caption{Comparing IRMAE against AEs with different latent dimension. Performed on CelebA dataset. IRMAE uses $l=4$ throughout the experiment. Results are listed in FID score.}
    \begin{tabular}{c|c|c|c|c|c}
\toprule
Latent dimension	&	32 &	64 &	128 &	256 & 512\\ 
\midrule
IRMAE ($l=4$)	&		81.6 &	64.6	& 47.6 & 42.7 &	42.0 \\
\midrule
AE		&	78.2 & 60.1 & 46.0 & 45.4 & 53.9 \\
\bottomrule
\end{tabular}
\label{tbl:dim}
\end{table}

\subsection{t-SNE visualization}

\begin{figure}[t!]
\centering
\settoheight{\tempdima}{\includegraphics[width=.45\linewidth]{example-image-a}}
\begin{tabular}{@{}c@{ }c@{ }c@{ }}
& multivariate
Gaussian  & GMM \\  
\rowname{AE}&
\includegraphics[width=0.45\textwidth]{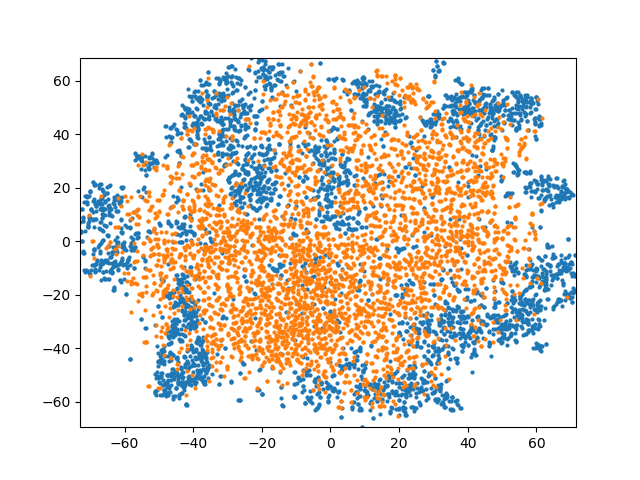} &
\includegraphics[width=0.45\textwidth]{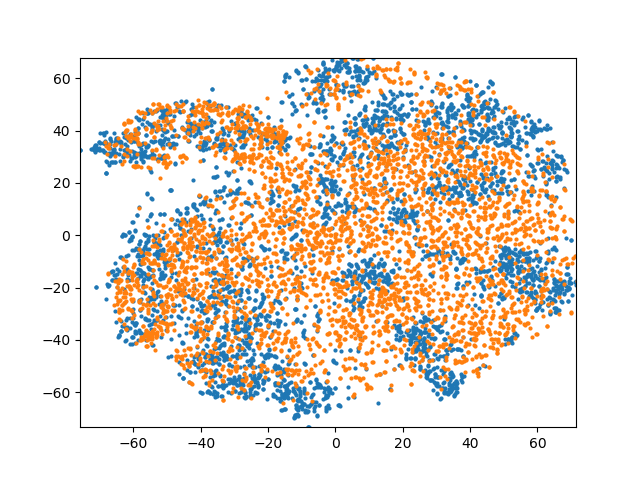} \\
\rowname{VAE}&
\includegraphics[width=0.45\textwidth]{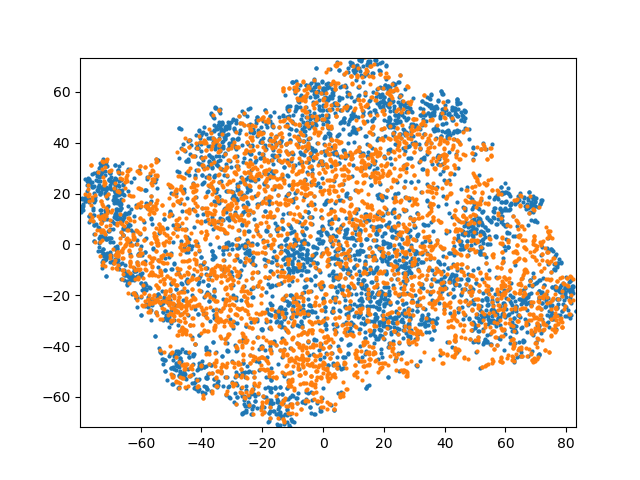}&
\includegraphics[width=0.45\textwidth]{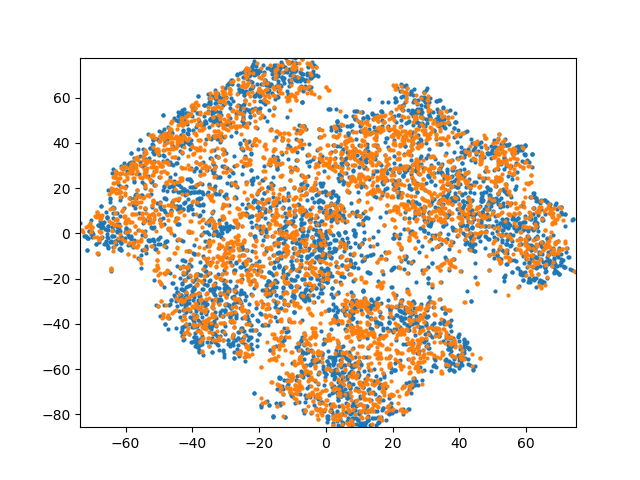} \\
\rowname{IRMAE}&
\includegraphics[width=0.45\textwidth]{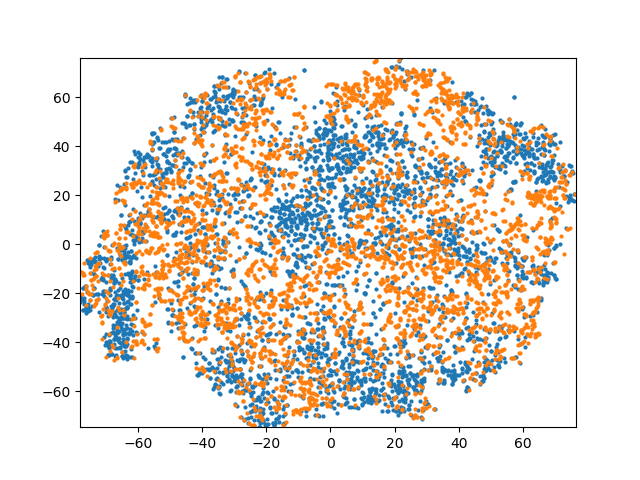} &
\includegraphics[width=0.45\textwidth]{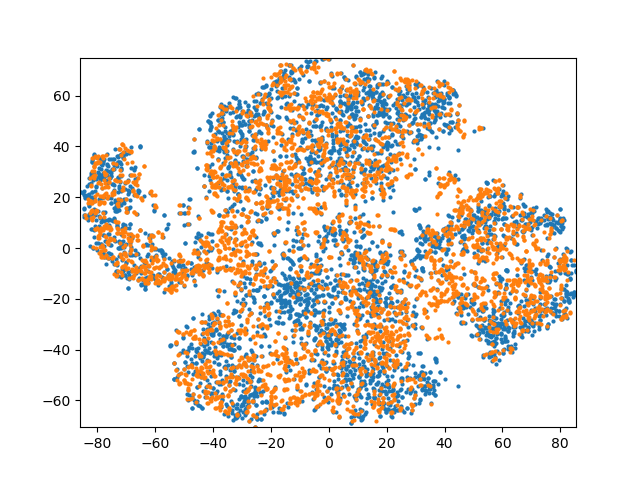}\\
\end{tabular}
\caption{t-SNE visualization on MNIST images. Blue points represent the test set data point. Orange points represent the sampled images.}
\label{fig:tsne}
\end{figure}

We visualize the density of the sampled MNIST images by each model in Figure \ref{fig:tsne} using t-SNE \cite{Maaten2008VisualizingDU}. Blue points represent the original data point, and the orange points represent the sampled ones. We compare IRMAE against an AE and a VAE. It's desirable that two point-clouds overlap. IRMAE demonstrates a comparable performance to VAE and a superior performance to AE.

\end{document}